\documentclass[11pt]{article}

\usepackage[]{acl}

\usepackage{times}
\usepackage{latexsym}
\usepackage{enumitem}
\usepackage{algorithm}
\usepackage{algorithmicx}
\usepackage{times}
\usepackage{latexsym}
\usepackage{graphicx} 
\usepackage{subfigure}
\usepackage{algpseudocode}
\usepackage{amsmath}
\usepackage{multirow}
\usepackage{booktabs}
\usepackage{color}
\usepackage{url}
\usepackage{array}
\usepackage{makecell}
\usepackage[normalem]{ulem}
\useunder{\uline}{\ul}{}
\usepackage[T1]{fontenc}

\usepackage[utf8]{inputenc}

\usepackage{microtype}
\newif\ifshowcomment
    \showcommenttrue

\ifshowcomment
    \newcommand{\fan}[1]{\textcolor{purple}{[{fan: #1}]}}
    \newcommand{\yang}[1]{\textcolor{blue}{[yang: #1]}}
\else
    \newcommand{\todo}[1]{}
    \newcommand{\fan}[1]{}
    \newcommand{\yang}[1]{}
    \newcommand{\focus}[1]{}
\fi

\title{Bridge the Gap Between CV and NLP!  \\
A Gradient-based Textual Adversarial Attack Framework
}

\author{
  Lifan Yuan$^{1}$\thanks{\ \ Equal contribution. Work done during internship at CCIIP lab.},
  Yichi Zhang$^{1}$\footnotemark[1],
  Yangyi Chen$^{2}$,
  Wei Wei$^{1}$\thanks{\ \ Corresponding author}
\\ 
$^{1}$Cognitive Computing and Intelligent Information Processing Laboratory, \\ School of Computer  Science and Technology,
Huazhong University of Science and Technology
 \\
$^{2}$University of Illinois Urbana-Champaign\\
{ \{lievanyuan173, phantivia, yangyichen6666\}@gmail.com}\\
{ weiw@hust.edu.cn}
}

\begin{document}
\maketitle

\begin{abstract}
Despite recent success on various tasks, deep learning techniques still perform poorly on adversarial examples with small perturbations. 
While optimization-based methods for adversarial attacks are well-explored in the field of computer vision, it is impractical to directly apply them in natural language processing due to the discrete nature of the text. 
To address the problem, we propose a unified framework to extend the existing optimization-based adversarial attack methods in the vision domain to craft textual adversarial samples. In this framework, continuously optimized perturbations are added to the embedding layer and amplified in the forward propagation process. Then the final perturbed latent representations are decoded with a masked language model head to obtain potential adversarial samples. 
In this paper, we instantiate our framework with an attack algorithm named \textbf{T}extual \textbf{P}rojected \textbf{G}radient \textbf{D}escent (\textbf{T-PGD}). 
We find our algorithm effective even using proxy gradient information.
Therefore, we perform the more challenging transfer black-box attack and conduct comprehensive experiments to evaluate our attack algorithm with several models on three benchmark datasets. Experimental results demonstrate that our method achieves overall better performance and produces more fluent and grammatical adversarial samples compared to strong baseline methods. The code and data are available at \url{https://github.com/Phantivia/T-PGD}. 
\end{abstract}

\section{Introduction}

\begin{figure}[h]
    \centering
    \includegraphics[ trim=180 150 180 130,clip, width=\linewidth]{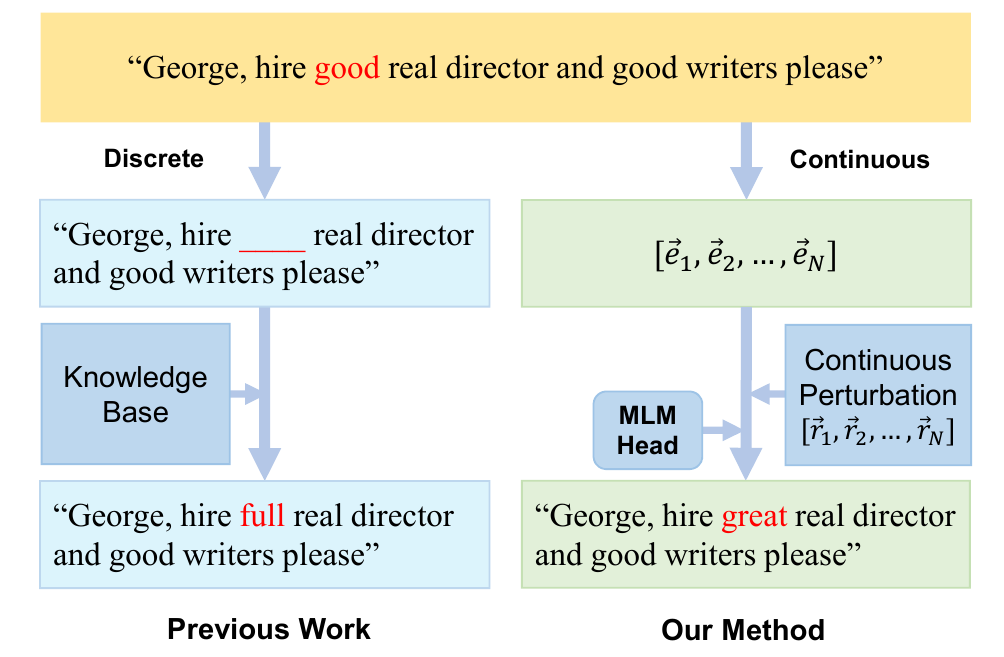}
    \caption{Comparison of our method with previous discrete substitution-based methods.
    }
    \label{fig:tmp}
\end{figure}

\looseness=-1
Despite great success in real-world applications, deep neural networks (DNNs) are still vulnerable to adversarial samples, which are crafted by adding small and human-imperceptible perturbations to the inputs and can change the prediction label of the victim model \citep{szegedy2014intriguing, goodfellow2015explaining}. 

In the field of computer vision (CV), numerous adversarial attack methods have been proposed to evaluate the robustness of DNNs \citep{papernot2016, madry2019deep}, and corresponding defense methods are also well-explored \citep{papernot2016distillation, Ross2018regularizing}. 
Adversarial attacks on images are defined as an optimization problem of maximizing the loss function of the model on specific samples, which can be approximated by gradient ascent algorithms. 

\looseness=-1
However, the textual adversarial attack is more challenging due to the discrete and non-differentiable nature of the text space. In Natural Language Processing (NLP), the methods that directly employ the gradients to optimize adversarial samples are not applicable in either the white-box or black-box settings, since they cannot obtain valid discrete texts.  
For this reason, most works in NLP explore some heuristic methods to produce discrete perturbations, such as manipulating the most important words in the text using corpus knowledge or contextualized information \citep{ren-etal-2019-generating, zang-etal-2020-word, li-etal-2020-bert-attack}. 
Besides, there are some practices of textual adversarial attacks that employ gradients for first-order approximation to find optimal candidates in vocabulary for word substitution, but the one-off search is less effective and can violate the local linearization assumption \citep{cheng-etal-2019-robust, behjati2019universal, xu2020TextTricker}.

To bridge this gap, we propose a general framework to adapt the existing optimization-based adversarial attack methods to NLP (See Figure~\ref{fig:tmp}). 
Essentially, we succeed in obtaining high-quality adversarial samples from the perturbed embedding space. 
Specifically, we employ gradients to produce perturbations on token embeddings rather than on the original text, thus transforming the problem of searching for adversarial samples in the discrete text space into searching in the continuous and differentiable embedding space. 
This provides the basis for applying adversarial attack methods investigated in CV to craft textual adversarial samples.
In this paper, we adapt the gradient-based algorithm PGD \citep{madry2019deep} within our framework to perform textual adversarial attacks, denoted as \textbf{T-PGD}.
Considering that in practical scenarios attackers may not hold the gradient information of the victim model, we explore the possibility of conducting a decision-based transfer attack.
To this end, besides the true victim model, we have another model dubbed the local proxy model in the attack process. \textbf{Gradient information comes from the local proxy model} and only the decision of the victim model can be accessed. 

\looseness=-1
Then the perturbed latent representations should be transferred back to the discrete text.
Although there have been some works exploring the feasibility of directly perturbing token embeddings \citep{sato2018interpretable, cheng-etal-2019-robust, behjati2019universal}, they simply use the first-order approximation of the gradient to select candidate words from vocabulary, which might break the local linearization hypothesis.
However, recent work finds that the mask language modeling (MLM) head can reconstruct input sentences from their hidden states with high accuracy, even after models have been fine-tuned on specific tasks \citep{kao2021berts}. 
Inspired by this, we employ an MLM head to decode the perturbed latent representations. With the extensive linguistic knowledge of MLM-head, the coherence and grammaticality of adversarial samples can be guaranteed.

\looseness=-1
We conduct comprehensive experiments to evaluate the effectiveness of our method by performing transfer black-box adversarial attacks, where only the final decisions of victim models are accessible, against three victim models on three benchmark datasets.
Experimental results demonstrate the effectiveness of our framework and T-PGD algorithm, with a higher attack success rate and more fluent and grammatical adversarial examples produced.

\looseness=-1
To summarize, the main contributions of this paper are as follows: (1) We propose a general textual adversarial attack framework facilitating NLP researchers to produce adversarial texts using optimization-based methods, bridging the gap between CV and NLP in the study of adversarial attacks. (2) Based on the framework, we propose an effective adversarial transfer attack method called T-PGD, handling the challenge of decision-based black-box attack, which is rarely investigated in NLP.

\section{Related Work}
\subsection{Adversarial Attack in CV}

In the field of computer vision, adding a small amount of perturbations to input images to mislead the classifier is possible \citep{szegedy2014intriguing} 
Based on this observation, various adversarial attack methods have been explored. FGSM \citep{goodfellow2015explaining} crafts adversarial samples using the gradient of the model's loss function to the input images. 
BIM \citep{kurakin2017adversarial} straightforwardly extends FGSM, iteratively applying adversarial perturbations multiple times with a smaller step size. MIM~\citep{dong2018boosting} exploits momentum when updating inputs, obtaining adversary samples with superior quality. 
PGD \citep{madry2019deep} employs uniform random noise as initialization. 
Both MIM and PGD are variants of BIM.  

Although well explored in CV, these methods are not directly transferable to NLP due to the discrete nature of the text.
A recent work GBDA \citep{guo-etal-2021-gradient} generates adversarial samples by searching an adversarial distribution, optimizing with a gradient-based algorithm that has been previously used in image adversarial attacks \citep{carlini2017towards}.  In this paper, we propose a general framework enabling the application of adversarial attacks in CV to text without many adaptions.

\subsection{Adversarial Attack in NLP}
Existing textual attacks can be roughly categorized into white-box and black-box attacks according to the accessibility to the victim models. 

\textbf{White-box attack} methods, also known as gradient-based attack methods, assume that the attacker has full knowledge of the victim models, including model structures and all parameters. There are few application scenarios of white-box attacks in real-world situations, so most white-box attack models are explored to reveal the weakness of victim models, including universal adversarial triggers \citep{wallace-etal-2019-universal}, and fast gradient sign inspired methods \citep{ebrahimi-etal-2018-hotflip,  papernot2016gradient}.

\looseness=-1
\textbf{Black-box attack} models can be further divided into two different attack settings, i.e. score-based and decision-based. 
The first one assumes the attacker can obtain the decisions and corresponding confidence scores from victim models. 
Most research works on black-box attacks focus on this setting, exploring different word substitution methods and search algorithms to reduce the victim models' confidence scores \cite{jin2020textfooler, ren-etal-2019-generating, zang-etal-2020-word, li-etal-2020-bert-attack, alzantot-etal-2018-generating}. 
The other attack setting assumes the attackers can only obtain decisions from victim models, which is more challenging and less studied. 
\citet{Maheshwary2021generating} first substitutes some words in the input sentences to flip the labels and then conducts a search based on a genetic algorithm, expecting to find the most semantic preserved adversarial samples.
\citet{chen2021multi} propose a learnable attack agent trained by imitation learning to perform a decision-based attack. Some works also explore sentence-level transformation, including syntax \citep{iyyer-etal-2018-adversarial} and text style \citep{qi2021mind}, to launch attacks. In this work, we consider the latter setting and show that even with less information, our decision-based attack can still be as effective as score-based ones.

\section{Framework}
\label{Framework}
\looseness=-1
In this section, we first present an overview of our framework, and next, we will give the details of how to add continuous perturbations and reconstruct the text.

\begin{figure}[thb]
    \centering
    \includegraphics[trim=215 180 265 115,clip, width=\linewidth]{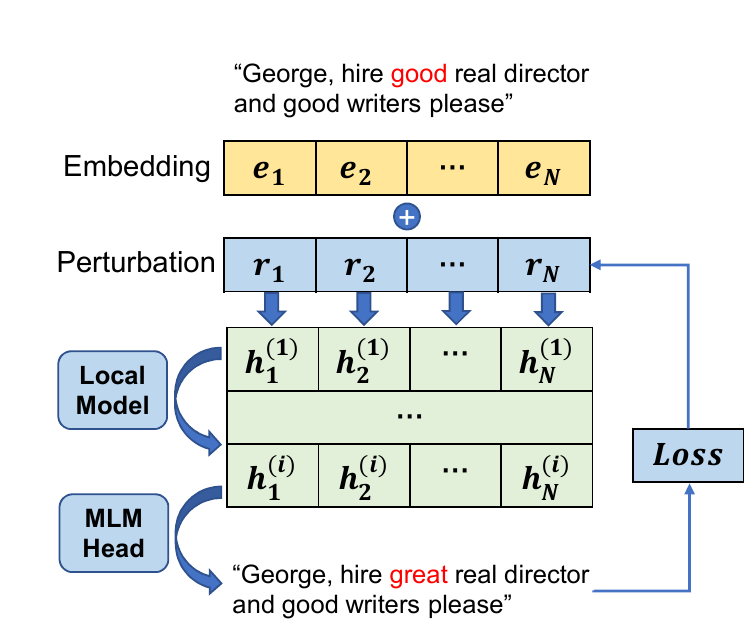}
    \caption{Overview of our framework. Continuous perturbations ($r_i$) are calculated as gradients of the loss function with respect to token embeddings. The MLM head is employed to decode the perturbed hidden states to obtain potential adversarial samples. 
    }

    \label{fig:architecture}
\end{figure}

\subsection{Overview}
\label{overview}

We have two models in the perturbation generation process: (1) a local proxy model which provides gradient information to optimize the adversarial samples, and (2) the true victim model that the attacker attempts to deceive. 
Specifically, a proxy BERT model fine-tuned on the attacker's local dataset encodes each discrete text instance into continuous token embeddings and then adds continuous perturbation to it. The perturbation would be iteratively optimized using the gradient of the proxy model, according to the prediction output of the victim model.
After perturbation, an MLM head will decode the perturbed latent representation to generate candidate adversarial samples.
The overview of the framework is shown in Figure~\ref{fig:architecture}.

With the help of our proposed framework, it is feasible to perform textual adversarial attacks with various gradient-based methods in CV.
In this paper, we examine PGD \citep{madry2019deep} as a case (See Section~\ref{approach}).

\subsection{Notation}
We denote each sample as $\left({x \in \mathcal{X},y \in \mathcal{Y}} \right)$, where $x$ denotes the input text,  $y$ denotes its corresponding label. 
In particular, the embeddings of $x$ is $e$, the hidden state is $h$, and final prediction is $\hat y$. The local neural network is implied by a mapping function $f$, which consists of three components, $f_0$, $f_1$, and $f_2$, holding:

\begin{equation}
    f\left( x \right) = {f_2}\left( {{f_1}\left( {{f_0}\left( x \right)} \right)} \right),
\end{equation}
where $f_0$ is the embedding layer, $f_1$ denotes the hidden layers from the first layer to $m$-th layer,  
and $f_2$ denotes the rest of the neural network. 
Then the forward propagation process can be described as:
\begin{eqnarray}
\begin{split}
e = {f_0}\left( x \right), \label{embed} h = {f_1}\left( e \right), \label{hidden}  \hat y = {f_2}\left( h \right)
\end{split}
\end{eqnarray}

\subsection{Latent-space Perturbation}
\looseness=-1
Previous work has shown that the latent representations of transformer-based pre-trained language models are effective in providing semantic and syntactic features \citep{clark2019bert, jawahar2019bert}, and thus we use a local BERT model fine-tuned on our local dataset as the encoder for our framework.

For each text input, we first calculate the task-specific loss in the forward propagation process, 
and then perform backward propagation to obtain the gradients of the loss with respect to the token embeddings of the input text. 
The generated gradients are viewed as the information for updating the perturbations added to the token embeddings, which can be obtained by solving an optimization problem as follows:

\begin{equation}
\label{optimization}
    \delta = \mathop {\arg \max }\limits_{\delta:{{\left\| \delta \right\|}_2} \le \varepsilon }\mathcal L\left( {f_2\left(f_1\left( f_0\left(x\right) + \delta\right)\right),y} \right),
\end{equation}
where $\delta$ is the perturbation and  $\mathcal L\left(\cdot  \right)$ is the loss function.

The closed-form solution to the optimization problem is hard to directly obtain \citep{goodfellow2015explaining}, which is thus relaxed to obtain an approximate solution.
For example, various methods in CV usually linearize the loss function with gradient information to approximate the perturbations $\delta$ \citep{goodfellow2015explaining, kurakin2017adversarial, madry2019deep}.

In NLP, most existing gradient-based methods commonly employ first-order approximation to obtain substitution words \citep{cheng-etal-2019-robust, behjati2019universal, xu2020TextTricker}. 
However, these one-off approaches may result in large step size perturbations, violating the hypothesis of local linearization (See Figure~\ref{fig:decode}).
To ensure the local linearization hypothesis, we consider adjusting the continuous perturbations added to the token embeddings with a minor change at each step, 
and then iteratively update the token embeddings of the input instance with the perturbations until generating a meaningful adversarial sample for attacking.

\subsection{Reconstruction}
\looseness=-1
Using continuous perturbations, we need to reconstruct the meaningful adversarial text from the optimized token embeddings.
The MLM-head is observed to be able to reconstruct input sentences from hidden states in middle layers with high accuracy, even after models have been fine-tuned on specific tasks \citep{kao2021berts}. 
Specifically, MLM-head is a pre-trained $H \times V$ linear layer, where  $H$  is the size of hidden states and  $V$  is the size of the vocabulary. Given continuous input hidden states $h$, it can predict token IDs by $t=h A^{T}+b$, where $A$ and $b$ are tuned parameters. The IDs can later be decoded into texts by the tokenizer using a predefined ID-token mapping.
Inspired by this, we adopt the MLM head as the decoder for two reasons: 1) MLM-head is capable of interpreting any representation embeddings in the hidden space, which is crucial to search adversarial examples continuously; 2) MLM-head has been fully trained during the pre-trained stage so it acquires linguistic knowledge together with the language model and can reconstruct sentences considering the contextual information.

\begin{figure}[t]
    \centering
    \includegraphics[trim=100 110 100 80,clip, width=\linewidth]{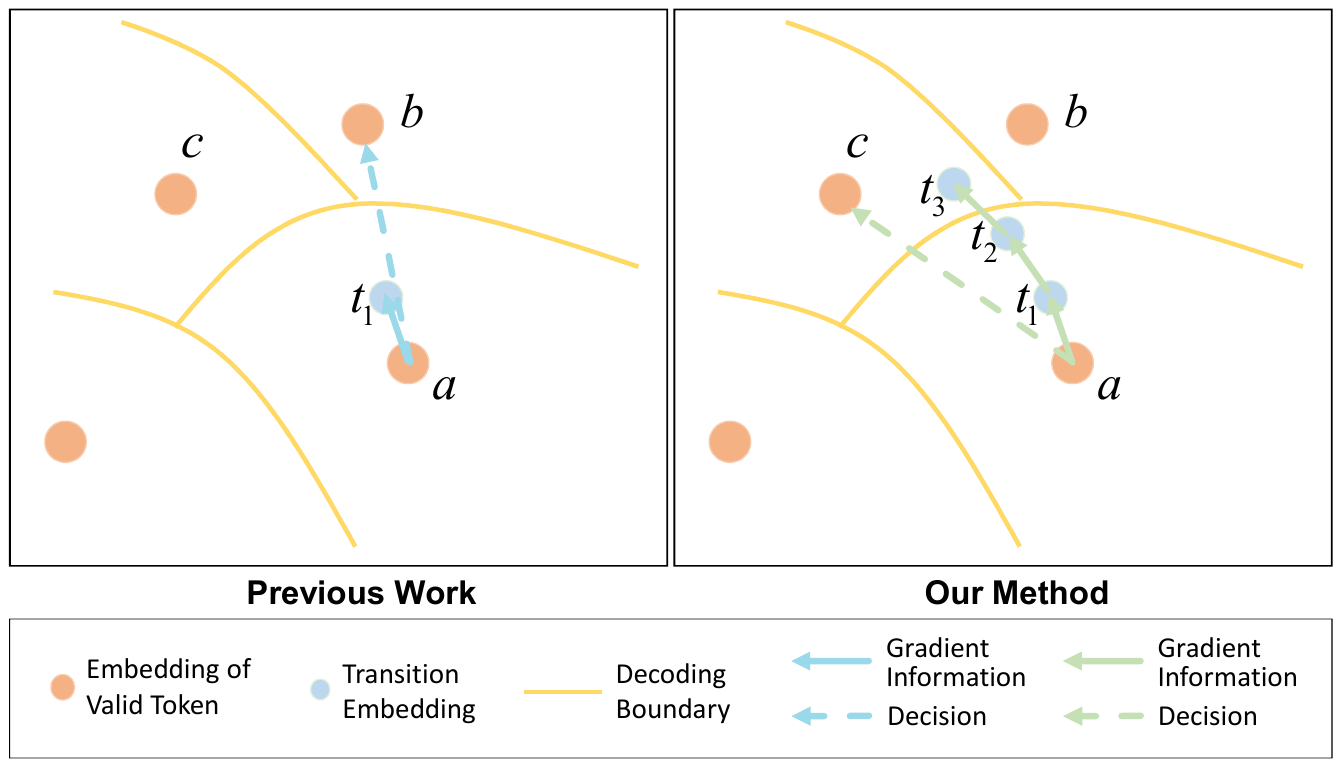}
    \caption{The process of searching for the substitute token of the original instance $a$ in the hidden space. 
    In this case, the one-off attack models are prone to select token $b$ after one-step perturbation (left), while our iterative perturbation-based method is more likely to find the optimal solution token $c$ (right). }
    
    \label{fig:decode}
\end{figure}

Without loss of generality, we take an example in Figure \ref{fig:decode} to illustrate the discrepancy between the one-off-based attack models and our proposed iterative-attack-based model. One-off attack models are prone to choose the token $b$ to serve as the substitute of token $a$ because $\cos (\overrightarrow {at_1} ,\overrightarrow {ab} ) < \cos (\overrightarrow {at_1} ,\overrightarrow {ac} )$.
However, in our framework, the one-step perturbation $\overrightarrow {at_1}$ does not cross the decoding boundary, and thus the decoding results remain unchanged if only using one-step perturbation. 
Based on the iterative search, the perturbations can be accumulated to the extent to cross the decision boundary and reach the transition point $t_3$, which will be decoded as the optimal solution $c$.
Then $a$ is replaced by $c$ to obtain the adversarial sample to query the victim model for its decision. 
If this adversarial sample fails to fool the victim model, we start the next searching iteration from the current perturbed token embedding, i.e. $t_3$ in Figure \ref{fig:decode}, 
but not from the embedding of the decoded token $c$.
By exploiting virtual embeddings as transition points, this iterative attack framework can preserve accumulated gradient information and avoid breaking local linearization assumptions.

\section{Method}
\label{approach}
\looseness=-1


\subsection{T-PGD Algorithm}

We instantiate our framework with PGD \citep{madry2019deep} algorithm, and name our attack model as \textbf{T}extual-\textbf{PGD} (\textbf{T-PGD}). The algorithm flow of T-PGD is shown in  Algorithm 1.

 \begin{algorithm}[!htbp]

 \caption{T-PGD}
 
     \begin{algorithmic}[1]
      \label{algo}
         \Require Original input $x$ sampled from $\mathcal{X}$
         \Ensure Adversary of $x$
         \State Randomly mask one word in $x$
         
         \State $AdvList = $ []
         \State $BestSim = 0 $
         \For{ $j \in [1, \ldots, MaxIter]$ }

            \State $e_0$ = $f_0\left(x\right)$ 
            \State $\delta_0 = \frac{1}{{ {{N_{{e_{0}}}}} }}Uniform\left( { - \varepsilon ,\varepsilon } \right)$
            
             \For{$i \in [1, \ldots, MaxStep] $}
                 
                 \State ${e_{i}} = {e_{i-1}} + {\delta _{i-1}}$
                 \State $h_{i} = {f_1}(e_{i})$  
                 \State $Adv_{i} = Dec(h_{i})$
                 \State $ Sim = USE\left(Adv_i, x\right)$
                 \If {$Adv_{i}$ \textbf{not in} $AdvList$ \textbf{and} $Sim > BestSim$} 
                     \State Append $Adv_{i}$ to $AdvList$
                     \State $BestSim = Sim$
                     \State Query victim model with $Adv_{i}$
                     \If{attack succeed \textbf{and} $Sim > Threshold$ \textbf{and}  no antonyms}
                         \State \Return $Adv_{i}$
                     \EndIf
                 \EndIf
                 \State ${g_{adv}} = {\nabla _{\delta_{i-1}} }\mathcal{L}\left( f_2\left(h_i\right) , y\right)$
                 \State ${\delta _{i}} = Proj_{{\left\| \delta  \right\|_F} \le \varepsilon }\left( {{\delta _{i-1}} + \alpha \frac{g_{adv}}{{\left\| {{g_{adv}}} \right\|}_F}} \right)$ 
             \EndFor
             
\EndFor
    \end{algorithmic}
\end{algorithm}

\looseness=-1
To solve the optimization problem in Eq. (\ref{optimization}), we iteratively search for the optimal solution by adding the gradient-based perturbations to the token embeddings.
For each sample, we first pre-defined a maximum iteration of the searching process to avoid the infinite loop problem.
In each iteration, we first map input $x$ to the token embeddings and initialize the perturbation by sampling noise from a uniform distribution.
 In the ${i}$-th step, we obtain new embeddings $E_i$ by adding $\delta_{i-1}$, the perturbation generated in the last step,  to $e_{i-1}$. Then, $e_{i}$ will be forward propagated to obtain a hidden representation: $h_i=f_1\left(e_i\right)$. Next, the hidden states with perturbations are decoded for reconstructing the crafted adversarial samples, $Adv_{i} = Dec(h_{i})$,
where $Adv_{i}$ denotes the adversarial sample obtained in this step. We then compute the semantic similarity $Sim_i$ between $Adv_i$ and input $x$ using Universal Sentence Encoder (\textit{USE}) score \cite{cer-etal-2018-universal}.

We query the victim model only when $Adv_{i}$ satisfying: (1) it varies from all potential adversarial samples that have been queried before; (2) it is more similar to the original sentences, compared to previous potential adversarial samples.
If the attack succeeds and $Sim$ is higher than a hyperparameter $Threshold$,  then $Adv_{i}$ is considered as the final adversarial sample of the original input. Otherwise, $h_i$ will be forwarded to obtain the prediction of the local model with respect to the input $x$. We then compute the loss between the predicted label and the golden label $y$ and then calculate the gradient w.r.t. $\delta_{i}$, and update the perturbation for next step, with the following formula: 

\begin{equation}
\label{pgd}
\begin{split}
&{g_{adv}} = {\nabla _{\delta_{i-1}} }\mathcal{L}\left( f_2\left(h_i\right) , y\right)\\
&{\delta _{i}} = Proj_{{\left\| \delta  \right\|_F} \le \varepsilon }\left( {{\delta _{i-1}} + \alpha \frac{g_{adv}}{{\left\| {{g_{adv}}} \right\|}_F}} \right),
\\
\end{split}
\end{equation}
where $g_{adv}$ is the gradient of the loss with respect to the continuous perturbation $\delta_{i-1}$, $\alpha$ is the step size of $\delta_{i-1}$, and $i$ denotes the current iteration step. 
$Proj\left(\cdot\right)$ performs a re-initialization when $\delta$ reaches beyond the $\epsilon$-neighborhood of the original embedding.

\subsection{Heuristic Strategies}
\label{sec:heuristics}
\paragraph{Random Masking for Diversity.} 
To enhance the diversity of adversarial samples, we randomly mask one token in each input sentence to randomly initialize the search for a broader search scope. 
Specifically, we tokenize x to a list of tokens, $x_{token} = [x_0, ... , x_i, ..., x_n]$. 
Then we randomly select $i$-th index token using the uniform distribution and replace it with a special token \textit{[MASK]}. 
Next, the MLM-head-based decoder will predict the masked word according to its context, which will diversify the generated adversarial samples with semantically consistent consideration.
Then, these processed sentences are embedded into continuous token embeddings as aforementioned.

\begin{table*}[t]
\centering

\tiny
\resizebox{\textwidth}{!}{
\begin{tabular}{ccccccccc}

\hline
\rule{0pt}{5.5pt} Dataset & \#Class & Train & Test & Avg Len & BERT Acc & RoBERTa Acc & ALBERT Acc & XLNET Acc\\ \hline

SST-2   & 2        & 7K    & 1.8K & 16.5    & 89.9     & 94.2        & 92.8       & 94.38\\
MNLI    & 3        & 433K  & 10K  & 31.7    & 82.8     & 83.6        & 82.3       & 87.06\\ 
AG's News      & 4        & 30K   & 1.9K & 39.3    & 91.2     & 94.7        & 94.2       & 98.96\\ \hline
\end{tabular}
}
\vspace{-5pt}
\caption{Detailed information of datasets and original accuracy of victim models.}
\label{dataset}
\end{table*}

\paragraph{Input Reconstruction Loss.}
\looseness=-1
\label{trainingstrategy}
Intuitively, the quality of generated adversarial samples is largely affected by the reconstruction accuracy of the MLM-head-based decoder.
If failing to recover the original sentence even with no perturbations added, its capacity to generate fluent adversarial samples from perturbed hidden states might be limited. 
Therefore, the MLM-head-based decoder should be constrained with external constraints to ensure reconstruction accuracy, thus guaranteeing the quality of generated adversarial samples.
Note that the MLM-head has been pre-trained to precisely fill the masked word, which is also fitted to our task. Hence, to preserve the reconstruction performance of the MLM-head in optimization, we add the MLM loss as a regularization term to the loss function.
Specifically, the loss function used in Eq. \ref{pgd} consists of two components:
\begin{equation}
\label{loss}
\mathcal{L}(f(x), y)=\mathcal{L}_{1}(f(x), y)+\beta \mathcal{L}_{2}(f(x), y),
\end{equation} 
\looseness=-1
where ${{\mathcal L}_{1}}\left( {f(x),y} \right)$ is the original loss of the local model on specific tasks (e.g. cross-entropy loss in sentiment classification), 
${{\mathcal L}_{2}}\left( {f(x),y} \right)$ is the CE loss of the input reconstruction task, and ${{\beta}}$ is a weighting constant. 
Considering that we aim to reduce the reconstruction loss ${\mathcal L}_{2}$ while increasing $\mathcal L\left( {f(x),y}\right)$ along the gradient direction, $\beta$ should be negative. 
Taking two losses into account jointly, we adjust the perturbation searching target to successfully fool the victim models with fewer modifications.

\paragraph{Selection for Layer Index $m$.} The layer index $m$ is dataset-specific but victim-agnostic. This is because there is a trade-off between ASR and USE when decoding different layers (layer index  $\uparrow$, USE $\uparrow$, ASR $\downarrow$  ). Therefore, we determine the  $m$  by tuning the USE score on a sampled dataset. In practice, we sample 100 examples and adopt BERT as the victim to conduct pilot experiments. We compute the USE scores of decoding different layers. We then set a USE threshold $t$ = 0.8  and disregard layers which leads to a USE score lower than  $t$. Finally, we find the lowest USE among the rest of the layers and set  $m$  as the index of the corresponding layer. We set $h$ = 10,11, and 7 for SST-2, MNLI, and AG, respectively. 

\paragraph{Antonym Filtering.}
\citet{Li_2019} reports that semantically opposite words locate closely in their representation embeddings since antonyms usually appear in similar contexts. 
Therefore, we filter antonyms of original words using WordNet \citep{Fellbaum2010wordnet} to prevent invalid adversarial samples.

\begin{table*}[]
\centering
\footnotesize

\renewcommand{\arraystretch}{1.1}
\resizebox{\textwidth}{!}{

\begin{tabular}{m{1.5cm}<{\centering}|m{2cm}<{\centering}|cccc|cccc|cccc|cccc}
\toprule
\multirow{2}{*}{Dataset}   & \multirow{2}{*}{Model} & \multicolumn{4}{c|}{BERT}                                         & \multicolumn{4}{c|}{RoBERTa}                                      & \multicolumn{4}{c|}{ALBERT}                                       & \multicolumn{4}{c}{XLNet}                                         \\ \cmidrule(l){3-18} 
                           &                        & ASR\%          & USE           & $\Delta$I      & $\Delta$PPL     & ASR\%          & USE           & $\Delta$I      & $\Delta$PPL     & ASR\%          & USE           & $\Delta$I      & $\Delta$PPL     & ASR\%          & USE           & $\Delta$I      & $\Delta$PPL     \\ \midrule
\multirow{6}{*}{SST-2}     & PWWS                   & 75.12          & 0.83          & 0.29           & 533.86          & 77.03          & 0.82          & 0.41           & 837.7           & 72.00          & 0.82          & 0.40           & 531.85          & 77.26          & 0.83          & 5.18           & 744.47          \\
                           & Textfooler             & 85.36          & 0.81          & 0.33           & 480.14          & 87.28          & 0.82          & 0.32           & 924.09          & 72.68          & 0.79          & 0.25           & 706.83          & 89.17          & 0.82          & 0.28           & 540.88          \\
                           & PSO                    & 85.60          & 0.75          & \textbf{0.10}  & 501.12          & 85.50          & 0.74          & \textbf{0.09}  & 479.27          & 91.49          & 0.77          & \textbf{0.14}  & 397.77          & 87.02          & 0.76          & \textbf{0.10}  & 498.94          \\
                           & BERT-Attack            & 90.36          & 0.81          & 0.51           & 378.79          & 93.53          & {\ul 0.88}    & 0.45           & 387.95          & 92.43          & 0.79          & 0.81           & 348.37          & 97.26          & 0.84          & 0.55           & 383.90          \\
                           & GBDA                   & 57.19          & 0.64          & 0.42           & \textbf{186.21} & 58.05          & 0.64          & 0.22           & \textbf{27.45}  & 54.31          & 0.64          & 0.47           & \textbf{153.94} & 56.56          & 0.64          & 0.22           & \textbf{28.34}  \\
                           & TPGD                   & \textbf{97.00} & \textbf{0.92} & 0.62           & 343.65          & \textbf{94.75} & {\ul 0.89}    & 0.63           & 302.70          & \textbf{93.59} & \textbf{0.90} & 0.69           & 291.00          & \textbf{97.29} & \textbf{0.91} & 0.65           & 334.55          \\ \midrule
\multirow{6}{*}{MNLI}      & PWWS                   & 75.12          & 0.83          & 0.34           & 516.95          & 71.65          & 0.84          & 0.3            & 715.42          & 45.88          & 0.77          & 4.17           & 744.49          & 75.10          & 0.83          & 0.34           & 316.95          \\
                           & Textfooler             & 72.34          & 0.83          & 0.31           & 780.8           & 77.27          & 0.87          & 0.3            & 640.21          & 82.47          & 0.81          & 0.31           & 854.73          & 84.70          & 0.82          & 0.31           & 1781.96         \\
                           & PSO                    & 75.85          & 0.8           & 0.11           & 481.43          & 76.08          & 0.80          & 0.11           & 411.12          & 89.41          & 0.79          & 0.22           & 424.48          & 75.80          & 0.80          & 0.11           & 381.43          \\
                           & BERT-Attack            & 87.68          & 0.87          & 0.55           & 484.27          & 91.26          & 0.89          & 0.23           & 604.22          & 89.65          & 0.89          & 0.25           & 456.31          & 82.10          & 0.79          & 0.55           & 10956.63        \\
                           & GBDA                   & 61.28          & 0.67          & 0.08           & \textbf{265.38} & 59.31          & 0.67          & 0.12           & 316.18          & 62.65          & 0.67          & 0.10           & 288.37          & 59.70          & 0.67          & 0.10           & \textbf{250.75} \\
                           & TPGD                   & \textbf{93.96} & \textbf{0.92} & \textbf{-0.95} & 296.82          & \textbf{94.55} & \textbf{0.91} & \textbf{-0.97} & \textbf{261.62} & \textbf{94.65} & \textbf{0.93} & \textbf{-0.98} & \textbf{259.57} & \textbf{93.63} & \textbf{0.90} & \textbf{-0.33} & 504.34          \\ \midrule
\multirow{6}{*}{AG's News} & PWWS                   & 65.46          & {\ul 0.84}    & 0.65           & 394.28          & 54.70          & 0.84          & 0.82           & 491.48          & 48.53          & 0.84          & 4.71           & 476.81          & 61.00          & 0.82          & 0.78           & 474.31          \\
                           & Textfooler             & 88.71          & 0.81          & 0.61           & 454.13          & 78.25          & 0.82          & 0.59           & 372.9           & 73.21          & 0.84          & 1.32           & 367.66          & 84.90          & 0.80          & 0.55           & 491.87          \\
                           & PSO                    & 66.22          & 0.79          & 0.25           & 539.25          & 64.63          & 0.79          & 0.29           & 508.76          & 76.37          & 0.84          & 0.15           & 282.73          & 61.30          & 0.78          & 0.33           & 565.82          \\
                           & BERT-Attack            & 81.25          & {\ul 0.84}    & 0.48           & 431.47          & 82.58          & 0.85          & 0.07           & 307.74          & 91.28          & 0.81          & 2.52           & 289.52          & 91.50          & 0.86          & 0.46           & 240.63          \\
                           & GBDA                   & 77.66          & 0.69          & -0.16          & \textbf{85.69}  & 68.97          & 0.69          & -0.59          & \textbf{96.95}  & 66.67          & 0.73          & 0.20           & \textbf{54.91}  & 71.16          & 0.67          & \textbf{-0.39} & \textbf{109.49} \\
                           & TPGD                   & \textbf{94.47} & 0.75          & \textbf{-0.05} & 625.08          & \textbf{99.30} & \textbf{0.87} & \textbf{-1.42} & 285.12          & \textbf{99.24} & \textbf{0.87} & \textbf{-1.14} & 260.64          & \textbf{94.05} & \textbf{0.89} & -0.10          & 277.17          \\ \bottomrule
\end{tabular}
}
\caption{The results of automatic evaluation metrics on SST-2, MNLI, and AG's News. 
ASR denotes the attack success rate, \textit{USE} denotes the similarity of original and adversarial samples, $\Delta$I and $\Delta$PPL denotes the increase of grammar errors and perplexity after original texts are transformed into adversaries. We conduct Student t-tests to measure the significant difference. 
\textbf{Bold} numbers indicate significant advantage with p-value 0.05 as the threshold and \underline{underline} numbers mean no significant difference.}
\label{metrics}
\vspace{-10pt}
\end{table*}

\section{Experiments}
We conduct comprehensive experiments to evaluate our general framework and T-PGD algorithm on the task of sentiment analysis, natural language inference, and news classification.
We consider both automatic and human evaluations to analyze our method in terms of attack performance, semantic consistency, and grammaticality.

\subsection{Datasets and Victim Models}
For sentiment analysis, we choose SST-2 \citep{socher2013sst2}, a binary sentiment classification benchmark dataset. 
For natural language inference, we choose the mismatched MNLI \citep{williams2018mnli} dataset. 
For news classification, we choose AG's News \citep{zhang2015agnews} multi-classification datasets with four categories: World, Sports, Business, and Science/Technology.  We randomly sample 1,000 samples that models can classify correctly from the test set and perform adversarial attacks on those samples. 

\looseness=-1
For each dataset, we evaluate T-PGD by attacking BERT \citep{devlin2019bert}, RoBERTa \citep{liu2019roberta}, ALBERT \citep{lan2020albert} and XLNet \citep{yang2019xlnet} with a local fine-tuned BERT model to generate potential adversarial samples. 
Details of datasets and the original accuracy of victim models are listed in Table~\ref{dataset}.

\subsection{Experimental Setting}
\paragraph{Baseline Methods.}

We select four strong score-based attacks as baselines: (1) PWWS \citep{ren-etal-2019-generating}; (2) Textfooler \citep{jin2020textfooler}; (3) PSO \citep{zang-etal-2020-word}; (4) BERT-Attack \citep{li-etal-2020-bert-attack}.
Note that all of them require the confidence scores of victim models, while our model only assumes the decisions are available, which is more challenging. We also make a comparison with the decision-based GBDA \citep{guo-etal-2021-gradient}.

\vspace{-0.0981in}
\paragraph{Evaluation Metrics.}
\looseness=-1
We evaluate our method considering the attack success rate and adversarial sample quality. (1) Attack Success Rate (\textbf{ASR}) is the proportion of adversarial samples that successfully mislead victim models' predictions.
(2) Quality of adversarial samples is evaluated by two automatic metrics and human evaluation, including their semantic consistency,  grammaticality, and fluency. Specifically, we use Universal Sentence Encoder \citep{cer-etal-2018-universal} to compute the semantic similarity between the original text and the corresponding adversarial sample, Language-Tool\footnote{\url{https://github.com/jxmorris12/language_tool_python}} to calculate the increase of grammar errors in texts after being perturbed, and GPT-2 \citep{radford2019gpt2} to compute the increase of perplexity to measure fluency. We also conduct a human evaluation to measure the validity and quality of adversarial samples.


\subsection{Experimental Results}
The results of automatic evaluation metrics are listed in Table~\ref{metrics}.
\paragraph{Attack Performance.}
T-PGD consistently outperforms the strong score-based attack methods considering the attack success rate.
We attribute the success of our attack method to the more effective searching process following the guidance of the gradient information, which is verified in the ablation study (Section \ref{ablation_gradient}).



\paragraph{Adversarial Sample Quality.}
\looseness=-1
We observe that the quality of the adversarial samples generated by T-PGD increases with the text length.
Our adversarial samples yield overall higher \textit{USE} scores than baseline models, indicating that our method can manipulate adversarial samples more precisely with explicit gradient information. 
And although the grammatical performance of T-PGD is not the best on SST-2, which mostly contains shorter text (See Table~\ref{dataset}), MNLI and AG's News T-PGD produce the fewest grammatical errors and the lowest perplexity, since the embedding space of longer text is broader and has a better optimal solution. Finally, we attribute the overall high quality of our adversarial samples to the introduction of reconstruction loss, which is demonstrated in Section \ref{sec:decode}.

\subsection{Human Evaluations}
To further study the quality and validity of adversarial samples, we randomly selected 100 original SST-2 sentences and 100 adversarial samples from the SOTA baseline BERT-Attack and T-PGD respectively for human evaluation. 
Following \citep{li-etal-2020-bert-attack}, we shuffle the 300 samples and ask 3 independent human judges to evaluate the quality (300 samples per person). 
For semantic consistency evaluation, we ask humans to predict the labels of mixed texts. 
For grammar and fluency, human judges score from 1 to 5 on the above examples. All annotators have no knowledge about the source of the text, and all their evaluation results are averaged (shown in Table~\ref{HumanEvel}).
\begin{table}[]
\centering
\resizebox{0.45\textwidth}{!}{
\begin{tabular}{ccc}

\hline
Source    & Accuracy & Grammar \& Fluency \\ \hline
Original    & 0.92     & 4.63              \\ 
BERT-Attack & 0.48     & 3.41              \\ 
T-PGD & 0.68     & 3.52             \\ \hline
\end{tabular}
}
\caption{Human evaluation on SST-2 in terms of prediction accuracy, grammar correctness, and fluency.}
\vspace{-12pt}
\label{HumanEvel}
\end{table}

\paragraph{Semantic Consistency.} Since human judges have high accuracy on the original text, the prediction results on texts can be regarded as ground truth labels. 
Therefore, human accuracy can be a criterion for semantic consistency between original sentences and adversarial ones. 
From the results, human judges achieve 0.68 accuracies on adversarial samples crafted by T-PGD, significantly higher than the baseline method.
This result verifies that the adversarial samples crafted by T-PGD have a better semantic consistency.

\paragraph{Grammar and Fluency.} 
We can also conclude from Table~\ref{HumanEvel} that adversarial samples crafted by T-PGD have better quality compared to the baseline method considering the grammar and fluency, evaluated by human annotators. 
However, both BERT-Attack and T-PGD suffer a decline in grammatical correctness and fluency of adversarial text, leaving room for improvement in future research.

\section{Further Analysis}
\looseness=-1

\paragraph{Importance of Gradient Information.}
\label{ablation_gradient}
\looseness=-1
T-PGD employs the gradient of the proxy local BERT model to approximate the perturbations.
To verify the effectiveness of the gradient information, we conduct an ablation experiment on SST-2 by adding only random perturbations in the embedding space without exploiting the gradient information. 
In detail, we generate a Gaussian noise with the same mean and variance as our gradient-based perturbations.
The results in Table~\ref{tab:random} shows that without exploiting the direction of the gradient, the search in embedding space may deviate from the vicinity where the optimal and original points are located, reflected by the low ASR and USE score respectively.

\begin{table}[hbt]
\centering

\renewcommand{\arraystretch}{1.1}
\resizebox{0.45\textwidth}{!}{
\begin{tabular}{m{1.5cm}<{\centering}|m{1.2cm}<{\centering}m{1.2cm}<{\centering}|m{1.2cm}<{\centering}m{1.2cm}<{\centering}}
\hline
\multirow{2}{*}{Model} & \multicolumn{2}{c|}{T-PGD} & \multicolumn{2}{c}{Random} \\ \cline{2-5} 
                                & ASR          & \textit{USE}        & ASR          & \textit{USE}         \\ \hline
BERT                            & 97.00        & 0.92       & 47.48        & 0.79       \\
RoBERTa                         & 94.75        & 0.89       & 56.59        & 0.79       \\
ALBERT                          & 93.59        & 0.90       & 51.36        & 0.79       \\ 
XLNET                          & 97.29        & 0.91       & 49.94        & 0.84       \\ 
\hline

\end{tabular}
}
\vspace{-5pt}
\caption{Ablation results of gradient information on SST-2. 
\textit{Random} corresponds to adding random perturbations to the embeddings.}
\vspace{-5pt}
\label{tab:random}
\end{table}

\begin{figure}[h]
    \centering
    \vspace{-10pt}
    \includegraphics[trim=195 180 175 165,clip, width=\linewidth]{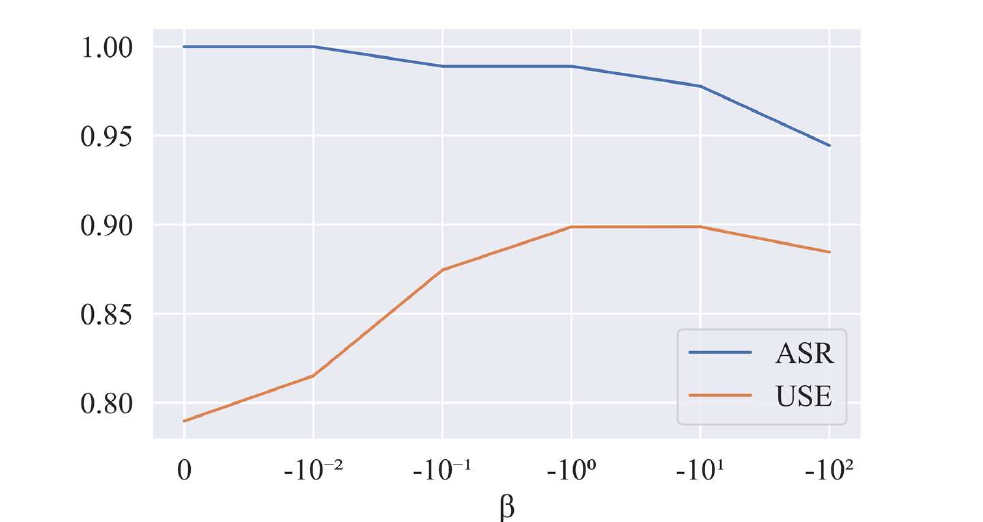}
    \vspace{-20pt}
    \caption{
    The trend of ASR and \textit{USE} with $\beta$ changing.}
    \label{fig:beta_bert}
    \vspace{-5pt}
\end{figure}

\paragraph{Importance of Reconstruction Task.} 
\label{sec:decode}
We show the importance of adding a reconstruction loss (${\mathcal L}_{2}$ in Eq.(~\ref{loss})) for generating more accurate reconstructions. 
We conduct an ablation study on SST-2. The results are shown in Table~\ref{tab:beta}. 
On all three victim models, the attack performances (ASR) improve significantly (up to 100) while the quality of adversarial samples deteriorates, with \textit{USE} score decreasing and grammar errors and perplexity increasing. 
This validates our claim that without reconstruction loss, the adversarial samples attempt to change the predictions of the model, ignoring whether the semantics is preserved and the linguistic quality is guaranteed. 
We further tune $\beta$ to study the trend of ASR and \textit{USE} score. 
Results on BERT are shown in Figure \ref{fig:beta_bert}. 
We observe that as the absolute value of $\beta$ increases, at the early stage ASR declines while \textit{USE} increases, suggesting that at first the effectiveness is sacrificed for sample quality; at the later stage ASR continues to decline and so does the \textit{USE}, showing that the reconstruction loss should not be over-weighted either.


\begin{table}[]
\centering
\normalsize
\renewcommand{\arraystretch}{1.2}
\resizebox{\linewidth}{!}{
\begin{tabular}{c|cccc|cccc}
\hline
\multirow{2}{*}{Victim} & \multicolumn{4}{c|}{T-PGD}       & \multicolumn{4}{c}{$\beta$=0}   \\ \cline{2-9} 
                        & ASR   & USE  & $\Delta$I & PPL    & ASR & USE  & $\Delta$I & PPL    \\ \hline
BERT                    & 97.00 & 0.92 & 0.62      & 343.65  & 100 & 0.79 & 1.45      & 875.64 \\
RoBERTa                 & 94.75 & 0.89 & 0.63      & 302.70  & 100 & 0.84 & 1.36      & 466.56 \\
ALBERT                  & 93.59 & 0.90 & 0.69      & 291.00   & 100 & 0.83 & 1.50      & 693.39 \\ 
XLNET                  & 97.29 & 0.91 & 0.65      & 334.55   & 99.42 & 0.83 & 1.24      & 623.23 \\ \hline
\end{tabular}
}
\vspace{-5pt}
\caption{Ablation results on the reconstruction loss. 
\textit{$\beta$=0} denotes the setting without the reconstruction loss.}
\label{tab:beta}
\vspace{-15pt}
\end{table}

\begin{figure}
    \centering
    \includegraphics[width=0.8\linewidth]{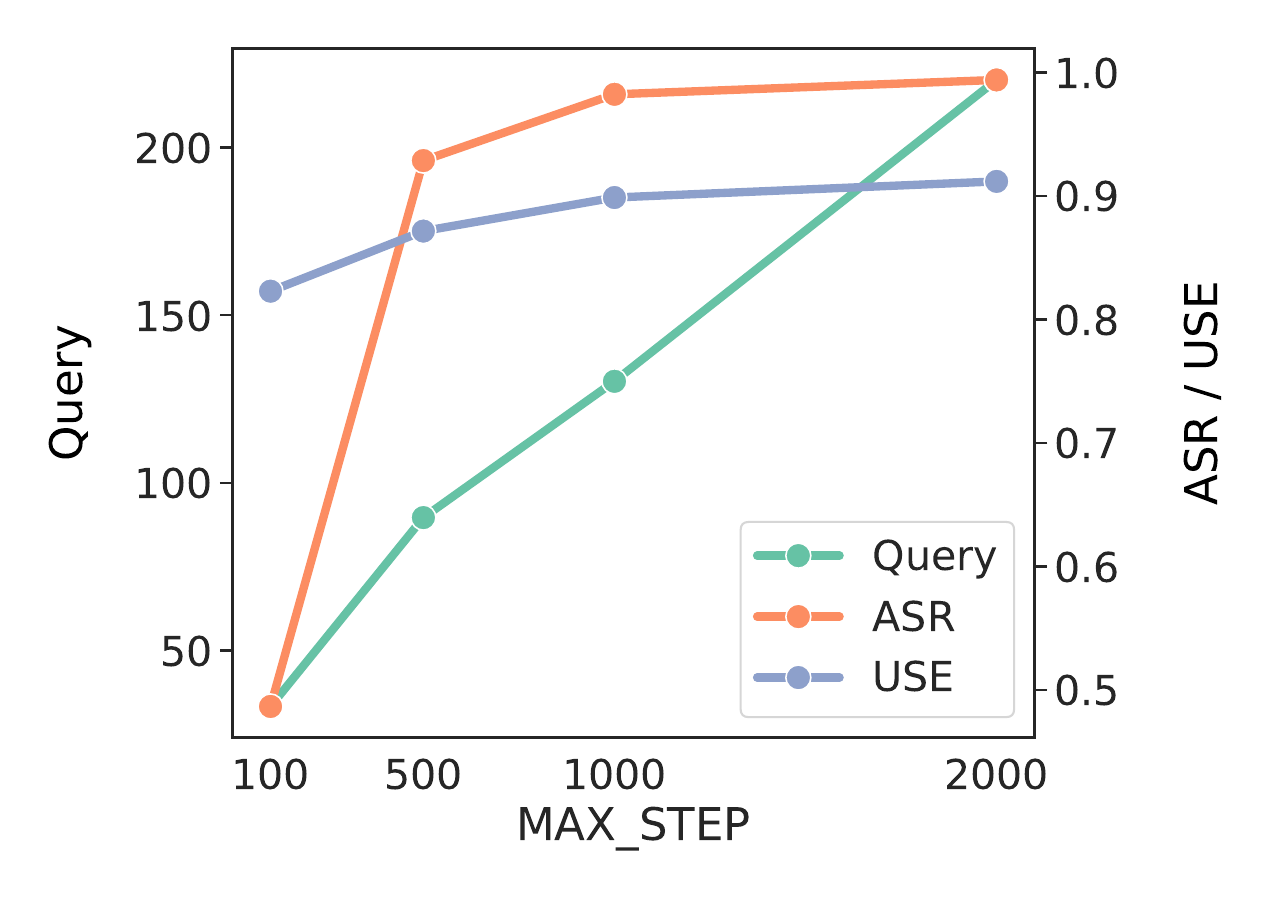}
    \vspace{-10pt}
    \caption{Trending of ASR, USE, and Query number with MAX\_STEP increasing. }
    \vspace{-8pt}
    \label{fig:query}
\end{figure}

\looseness=-1
\paragraph{Efficiency and Imperceptibility.} 
Despite T-PGD presenting impressive effectiveness in Table \ref{metrics}, it is also important to figure out if it is obtained by sacrificing efficiency and imperceptibility. Therefore, we examine the query number and perturbation rate by attacking XLNET on SST-2. Results are shown in Table \ref{tab:efficiency_imperceptibility}. We observe that T-PGD has the lowest perturbation rate, but the query number is relatively high. Hence, we conduct a more detailed experiment to set different MaxStep to track the trend of ASR, USE, and query number. As shown in Figure \ref{fig:query}, we can see that by fixing MaxStep to 500, TPGD can still perform a strong attack (ASR=\textit{89.27}) with a low query budget (Query=\textit{89.91}).
In conclusion, despite we require a relatively high query number to achieve the reported result, we can resort to an efficient version of TPGD which still achieves very competitive ASR. 

\begin{table}[htb]
\resizebox{\linewidth}{!}{
\begin{tabular}{l|cccc}
\hline
Attacker    & ASR   & USE  & Query   & Pert. (\%)  \\ \hline
PWWS        & 77.26 & 0.83 & 147.11  & 20.21                  \\
Textfooler  & 89.17 & 0.82 & 97.14   & 20.16                  \\
PSO         & 87.02 & 0.76 & 5113.83 & 15.96                  \\
BERT-Attack & 97.26 & 0.84 & \textbf{66.82}   & 23.83                  \\
GBDA        & 56.56 & 0.64 & 102.53  & 44.98                  \\
T-PGD       & \textbf{97.29} & \textbf{0.91} & 211.20  & \textbf{14.84}                  \\ \hline
\end{tabular}

}
\caption{Result of efficiency and imperceptibility on attacking XLNET-SST-2, where Pert. is the abbreviation for perturbation rate. Lower query number and perturbation rate indicate better efficiency and imperceptibility respectively.}
\label{tab:efficiency_imperceptibility}
\end{table}

\vspace{-13pt}
\paragraph{Transferability Across Models.}
\label{sec:Transferability}
We investigate the transferability of adversarial examples. 
We sample 1,000 samples from SST-2 and craft adversarial samples by T-PGD and baseline methods by attacking BERT. 
Then we test the attack success rate of these adversarial samples on RoBERTa to evaluate the transferability of adversarial samples.
As seen in Table~\ref{tab:trans}, adversarial samples crafted by T-PGD achieve the best transferability performance.
\begin{table}[htb]
\centering
\normalsize 
\renewcommand{\arraystretch}{1.5}
\resizebox{0.48\textwidth}{!}{

\begin{tabular}{c|ccccc}
\hline
 Method  & PWWS  & Textfooler   & PSO & BERT-Attack  & TPGD           \\ \hline
 Transfer ASR & 28.21 & 18.00 & 44.73      & 11.02    & \textbf{45.29} \\ \hline
\end{tabular}
}
\vspace{-3pt}
\caption{The ASR on SST-2 of attacking RoBERTa using adversarial samples crafted on attacking BERT.
}
\label{tab:trans}
\vspace{-20pt}
\end{table}

\paragraph{Transferability Across Training Datasets.}
\looseness=-1
We consider a more practical setting in which the attacker does not have the same downstream training dataset as the victim, i.e. the local proxy model is trained on a different dataset from the victim model.
To this end, we train a local proxy BERT model on another sentiment analysis dataset, IMDB or Amazon, and attack the victim model on SST-2. 
We compared the results with attacking with the local proxy model trained on the same dataset as the true victim model in Table~\ref{tab:distribution_shift}.
We can see that T-PGD can also achieve great attack performance in these practical circumstances, although slightly worse than training on the same dataset.


\begin{table}[htb]
\centering
\tiny
\renewcommand{\arraystretch}{1.15}
\resizebox{\linewidth}{!}{
\begin{tabular}{c|cccc}
\hline
Victim  & \multicolumn{4}{c}{BERT-SST-2}         \\ \hline
Dataset & ASR   & USE  & $\Delta$I & $\Delta$PPL \\ \hline
SST-2   & 97.00 & 0.92 & 0.62      & 343.65      \\
IMDB    & 93.30 & 0.90 & 0.70      & 204.18      \\ 
Amazon  & 96.40 & 0.91 & 1.00      & 388.93      \\
\hline
\end{tabular}
}
\caption{ Results of transferability across datasets. 
The local model is fine-tuned on SST-2, IMDB, and Amazon respectively.
}
\label{tab:distribution_shift}
\end{table}

\section{Conclusion and Future Work}
In this paper, we propose a general framework to facilitate generating discrete adversarial texts using optimization-based methods.
In our framework, the problem of searching textual adversarial samples in discrete text space is transformed into the continuous embedding space, where the perturbation can be optimized by gradient information, as explored in CV.
The perturbations in embeddings will be amplified in the forward propagation process, then decoded by an MLM head from the latent representations. 
We instantiate our framework with T-PGD, where the gradient comes from the local proxy model instead of the true victim model, i.e. T-PGD performs a decision-based black-box attack.
Experimental results show the superiority of our method in terms of attack performance and adversarial sample quality.

\looseness=-1
In the future, we will adopt other methods in CV with our framework.
Besides, we find that our framework can serve as a general optimization framework for discrete texts, and thus has the potential to provide solutions to other tasks like text generation. 
We will further explore this direction.

\section*{Limitations}
\looseness=-1
In experiments we only take PLMs into account because of their prevalence, hence the transferability to non-pretrained models is still unknown. However, due to the generality of PLMs, this can be a minor point in practical scenarios. 
Moreover, although we successfully transfer adversarial attack methods in CV to NLP using a unified framework, we only instantiate the framework with the PGD attack as an example. It would be interesting to transfer more attack methods in CV and conduct a comprehensive analysis of what methods can benefit NLP, aiming to have a deeper understanding of PLMs.

\section*{Ethical Consideration}
In this section, we discuss the potential broader impact and ethical considerations of our paper. 
\paragraph{Intended Use.} In this paper, we design a general framework to adapt existing gradient-based methods in CV to NLP, and further, propose a decision-based textual attack method with impressive performance. Our motivations are twofold. First, we attempt to introduce adversarial attack methods of CV to NLP, since image attack methods have been well-explored and proved to be effective, therefore helping these two fields better share research resources hence accelerating the research process on both sides. Second, we hope to find insights into the interpretability and robustness of current black-box DNNs from our study.

\paragraph{Potential Risk.} There is a possibility that our attack methods may be used maliciously to launch adversarial attacks against off-the-shelf commercial systems.
However, studies on adversarial attacks are still necessary since it is important for the research community to understand these powerful attack models before defending against these attacks.

\paragraph{Energy Saving. } We will public the settings of hyper-parameters of our method, to prevent people from conducting unnecessary tuning and help researchers quickly reproduce our results. We will also release the checkpoints including all victim models to avoid repeated energy costs.

\section*{Acknowledgement}
This work was supported in part by the National Natural Science Foundation of China under Grant No. 62276110, in part by CCF-AFSG Research Fund under Grant No.RF20210005, and in part by the fund of Joint Laboratory of HUST and Pingan Property \& Casualty Research (HPL). Thanks to Naixi Chen from SRU for providing hardware maintainance support for this work.
The authors would also like to thank the anonymous reviewers for their comments on improving the quality of this paper.

\bibliography{anthology,custom}
\bibliographystyle{acl_natbib}

\appendix
\newpage

\section{Adversarial Training}
We explore to enhance models' robustness against adversarial attacks through adversarial training on SST-2 with BERT.
Specifically, we first generate adversarial samples using the original training dataset.
Then we fine-tune the BERT model using the training dataset augmented with generated adversarial samples. 
We evaluate the model's  original accuracy on the test set and robustness against different adversarial attack methods. 
As seen in Table~\ref{tab:advTrain}, the model shows generally better robustness through adversarial training.
Besides, the accuracy on the test set is also improved from 89.90 to 90.48, which is different from previous textual adversarial attacks where accuracy is sacrificed for robustness \citep{ren-etal-2019-generating, zang-etal-2020-word}.

\begin{table}[H]
\centering

\resizebox{\linewidth}{!}{
\begin{tabular}{c|c|c|c|c|c}
\hline
Ori Acc & \multicolumn{5}{c}{89.90\%}                                    \\ \hline
Adv.T Acc & \multicolumn{5}{c}{\textbf{90.48\%}}                           \\ \hline
Method  & PWWS                  & Textfooler            & PSO                       & BERT-Attack       & T-PGD \\ \hline
Ori ASR & 69.94                 & \textbf{86.38}     & 82.03                     & 86.55             & 92.22     \\
Adv.T ASR & \textbf{66.78}   & 87.41                 & \textbf{73.34}         & \textbf{84.84} & \textbf{83.78}     \\ \hline
\end{tabular}
}

\caption{Results of adversarial training. \textit{Adv.T} denotes the adversarial training paradigm. 
}
\label{tab:advTrain}
\end{table}

\section{Ablation Study of Random Masking}
We conduct an ablation study of random masking. 
Our intuition is that random masking can broaden the searching scope of adversarial examples, and thus lead to diverse adversarial samples and higher attack success rate.
To prove this, we attack BERT on SST-2, with and without our random masking strategy. 
Result are shown in Table~\ref{tab:random_masking}.

\begin{table}[H]
\tiny
\centering
\renewcommand{\arraystretch}{1.1}
\resizebox{0.9\linewidth}{!}{
\begin{tabular}{c|cc|cc}
\hline
\multirow{2}{*}{Model} & \multicolumn{2}{c|}{w} & \multicolumn{2}{c}{w/o} \\ \cline{2-5} 
                       & ASR          & USE        & ASR           & USE         \\ \hline
BERT                   & 97.00        & 0.92       & 92.20         & 0.91        \\ \hline
\end{tabular}
}
\caption{Ablation results of random masking on SST-2 against BERT. }
\label{tab:random_masking}
\end{table}

\section{Case Study}
In Table \ref{tab:case_study}, we present some cases of our adversarial samples which successfully fooled XLNET.
\begin{table*}[t]
\centering
\begin{tabular}{l|l|p{12cm}}
\hline
Dataset                    & Type & Text                                                                                                                           \\ \hline
\multirow{4}{*}{SST-2}     & Ori  & the movie \textbf{bounces} all over \textbf{the} map.                                                                                            \\
                           & Adv  & the movie \textbf{bounce \&} all over \textbf{\&} map.                                                                                           \\ \cline{2-3} 
                           & Ori  & looks like a \textbf{high} school film project completed the day before it was due.                                                     \\
                           & Adv  & looks like a \textbf{unique} school film project completed the day before it was due.                                                   \\ \hline
\multirow{4}{*}{MNLI}      & Ori  & PREMISE: and he said , what 's going on ? HYPOTHESIS: he wanted to know what was going on \textbf{.}                                    \\
                           & Adv  & PREMISE: and he said , what 's going on ? HYPOTHESIS: he wanted to know what was going on \textbf{¡}                                    \\ \cline{2-3} 
                           & Ori  & PREMISE: they seem to have him on a primary radar . HYPOTHESIS: they \textbf{have} got him on a primary radar .                         \\
                           & Adv  & PREMISE: they seem to have him on a primary radar . HYPOTHESIS: they \textbf{finally} got him on a primary radar.                       \\ \hline
\multirow{4}{*}{AG's News} & Ori  & nortel lowers expectations nortel said \textbf{it} expects \textbf{revenue} for the third quarter to fall short of expectations .                \\
                           & Adv  & nortel lowers expectations nortel said , expects \textbf{income} for the third quarter to fall short of expectations .                  \\ \cline{2-3} 
                           & Ori  & \textbf{itunes} now selling band aid song \textbf{ipod owners} can download the band aid single after apple reaches agreement with the charity . \\
                           & Adv  & \textbf{the} now selling band aid song \textbf{dar norman} can \textbf{reach} the band aid single after apple reaches agreement with the charity.         \\ \hline
\end{tabular}
\caption{Cases of adversarial examples generated by T-PGD. The differences between original and adversarial texts are in \textbf{bold}.}
\label{tab:case_study}
\end{table*}

\end{document}